# A Comparative Evaluation of Structural Topic Models and BERTopic for Short, Open-Ended Survey Responses

Yan Jiang, Sihong Liu, Philip A. Fisher

Stanford Center on Early Childhood, Stanford University

## Author Note

Yan Jiang https://orcid.org/0000-0002-8825-8641

Sihong Liu https://orcid.org/0000-0002-5188-5334

Philip A. Fisher https://orcid.org/0000-0002-5399-7233

We have no known conflict of interest to disclose.

Correspondence concerning this article should be addressed to Yan Jiang, Stanford Center on Early Childhood, 520 Galvez Mall, CERAS 407, Stanford, CA 94305. Email: yannj@stanford.edu

**Abstract**

Topic modeling in applied psychology increasingly spans two methodological traditions: probabilistic bag-of-words models and newer embedding-based approaches. Yet many evaluations of these methods rely on longer and cleaner benchmark corpora, leaving less guidance for short, open-ended survey responses. This paper compares Structural Topic Models (STM), a probabilistic topic model, and BERTopic, an embedding-based model, for analyzing open-ended survey responses. We evaluated three STM conditions and five BERTopic conditions, varying typographical correction, stemming, embedding choice, and contextual augmentation, a strategy we introduced to provide additional semantic context for very short responses. Results indicate that BERTopic consistently produced higher topic coherence than STM, with contextual augmentation yielding the strongest performance gains. In contrast, higher-dimensional embeddings alone did not improve coherence and were associated with greater data loss. Qualitative evaluation showed that BERTopic generated more interpretable and stable topics, while STM topics were often broader and more mixed. However, STM provides stronger support for inferential covariate analysis, whereas BERTopic covariate comparisons are primarily descriptive. These findings suggest that STM and BERTopic offer complementary strengths. We conclude with practical guidance for selecting and combining topic modeling approaches in applied social science research.

# A Comparative Evaluation of Structural Topic Models and BERTopic for Short, Open-Ended Survey Responses

The rapid growth of digital text has created new opportunities for social scientists to study meaning, experience, and social conditions at scale (Roberts et al., 2014). In psychology and the broader social sciences, open-ended text responses are valuable because they capture participants' experiences and interpretations in their own words, offering insights that may not be available from numerical measures alone. As a result, computational analysis of open-ended survey responses has become increasingly common in psychological and social science research (Brickman et al., 2025; Feuerriegel et al., 2025).

Among these approaches, topic modeling has emerged as a widely used method for identifying latent themes in large collections of documents. Broadly, topic models uncover recurring patterns of word use and organize them into interpretable topics (Blei et al., 2003). Early probabilistic topic models, such as Latent Dirichlet Allocation (LDA; Blei et al., 2003), rely on a bag-of-words representation in which documents are treated as collections of words without considering word order or context. More recent advances in natural language processing have led to embedding-based methods that leverage pretrained language models to capture semantic relationships between words and documents (Sia et al., 2020). As embedding-based approaches have become increasingly prominent in natural language processing, there is a need to better understand how they can be applied in psychology and the broader social sciences alongside widely used probabilistic topic models.

Despite these developments, topic modeling methods are often evaluated on benchmark corpora, such as news articles or other relatively long and coherent documents, that differ from many forms of applied social science text (Grootendorst, 2022; Laureate et al., 2023). Open-ended survey responses, for example, are often short, uneven in length, and noisy. Some responses contain only one or two words, while others include spelling errors or fragmented phrases. These characteristics are common in applied research settings but are underrepresented in the corpora typically used to develop and evaluate topic modeling methods. As these methods are increasingly applied in practice, it is important to understand how topic models behave under such data conditions, and how different modeling choices and conditions affect performance.

In this article, we focus on Structural Topic Models (STM) and BERTopic as representative examples of these two modeling traditions. STM is a widely used probabilistic topic model in psychology and the broader social sciences because it supports covariate-based inference (Roberts et al., 2014). BERTopic is a widely used embedding-based approach that clusters semantically similar documents and generates topic representations from these clusters (Grootendorst, 2022). Rather than treating the comparison as a simple contest between models, we use it to examine how each approach responds to several analytic decisions relevant to short survey response data. These include established but debated choices, such as typographical correction, stemming (reducing words to their root form), and embedding model selection, as well as contextual augmentation, a strategy we introduce to provide additional semantic context for very short responses. Our goal is to clarify the strengths and constraints of each approach and to provide practical guidance for researchers selecting and applying topic modeling methods in applied social science settings.

We proceed by first describing the underlying assumptions, workflows, and applications of STM and BERTopic. We then introduce the dataset, preprocessing procedures, model specifications, and evaluation metrics used in this study. Next, we present the results of our comparative analyses across different model conditions. Finally, we discuss the implications of these findings and offer practical recommendations for future work.

**Structural Topic Models**

STM were introduced by Roberts et al. (2014) and have become widely used in the social sciences. STM belongs to the broader family of probabilistic topic models that build on Latent Dirichlet Allocation (LDA; Blei et al., 2003).

LDA is a generative probabilistic model that represents each document as a mixture of latent topics, where each topic is defined as a probability distribution over words (Blei et al., 2003). Figure 1 provides an intuitive geometric illustration of this structure, showing how documents can be represented as mixtures of topics and how topics correspond to distributions over words. In practice, the number of topics ($K$) is specified by the researcher prior to model estimation. Under this framework, documents are assumed to arise from a generative process in which a document-specific topic distribution is first drawn, and words are then generated from those topics (Figure 1). The model estimates both document-level topic mixtures and topic-level word distributions that best explain the observed corpus. As a result, each document can be characterized by multiple topics with different probabilities, a property often referred to as mixed membership (Grimmer & Stewart, 2013).

**Figure 1**

*Simplified Geometric Interpretation and Generative Process of Latent Dirichlet Allocation*

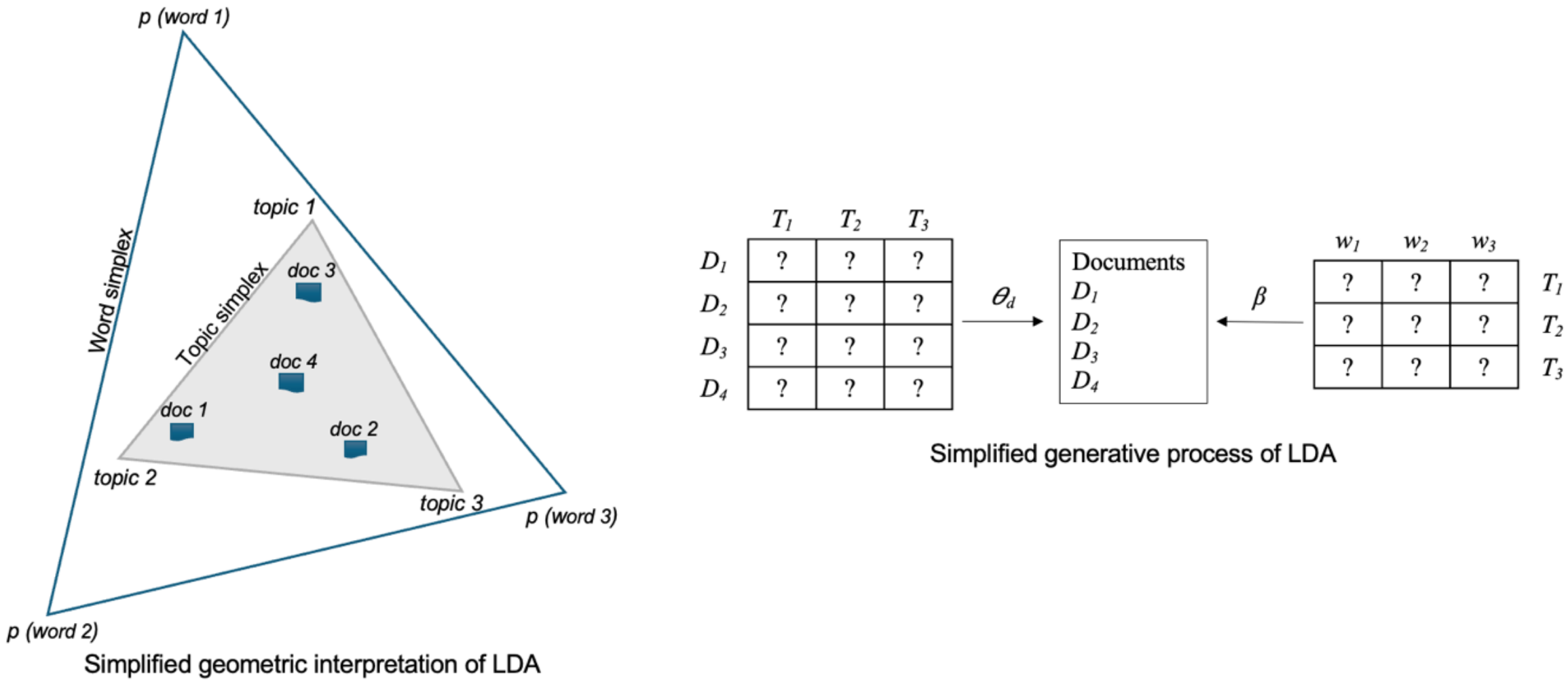

*Notes.* Based on frameworks presented in De La Hoz-M et al. (2022) and Roberts et al. (2019). The topic simplex (inner triangle) represents documents as mixtures of topics, while the word simplex (outer triangle) represents topics as distributions over words. In the generative process, each document is associated with a topic distribution ($\theta$), and each topic with a word distribution ($\beta$). Documents are generated by repeatedly selecting a topic from $\theta$ and a word from $\beta$, resulting in observed text as a mixture of topics.

LDA assumes that topics are independent and does not explicitly model correlations among them. Subsequent extensions, such as Correlated Topic Models (CTM; Blei & Lafferty, 2007), relax this assumption by allowing topics to be correlated. STM extends this line of work by incorporating covariates directly into the topic estimation process. Specifically, STM allows researchers to examine how topic prevalence and topic content vary as a function of covariates, such as group membership or time (Roberts et al., 2014). When no covariates are included, STM reduces to an implementation of CTM (Roberts et al., 2019).

Like LDA and related models, STM relies on the bag-of-words assumption, in which documents are treated as unordered collections of words (Blei et al., 2003). This implies that word order and syntax are not directly modeled. For example, sentences such as “a dog eats a donut” and “a donut eats a dog” would be treated as equivalent in terms of word counts. Similarly, bag-of-words representations may struggle to capture semantic similarity across different word choices (e.g., “low pay” versus “low wage”) and may not distinguish between words with multiple meanings (e.g., “bank” as a financial institution versus a riverbank).

Contextual differences can also be difficult to represent, as the same word may appear in different meanings across sentences. Table 1 summarizes the main features of STM and BERTopic.

**Table 1**

*Key Feature Comparison Between STM and BERTopic*

| Feature | STM | BERTopic |
|---|---|---|
| Broader tradition | Probabilistic topic modeling | Embedding-based topic modeling |
| Text representation | Bag-of-words representation | Document embeddings |
| Basis for topic formation | Primarily on word frequency distributions | Primarily on semantic similarity in embedding space |
| Model logic | Generative probabilistic model | Non-generative pipeline combining embeddings, dimensionality reduction, clustering, and c-TF-IDF |
| Topic membership | Mixed membership | Single membership |
| Number of topics | Specified by the researcher before estimation | Emerges from clustering |
| Covariate analysis | Incorporated during topic estimation, supporting inferential analysis | Conducted after topics are generated, mainly descriptive |
| Excluded data | Words may be removed during preprocessing | Documents that do not fit dense clusters may be assigned to an outlier category |

Despite these constraints, STM remains widely used and valuable in psychology and the broader social sciences for both empirical studies and meta-scientific analyses. For example, researchers have applied STM to analyze narrative data, such as personal accounts of identity-related experiences, and to examine how themes vary across demographic groups (Jacobson et al., 2024). Other studies have used STM to analyze written responses to understand behavioral patterns and group differences (Engstrom et al., 2025). In addition, STM has been used to map the structure of research fields over time by identifying trends in published literature (e.g., Wieczorek et al., 2021).

## BERTopic

BERTopic was developed by Grootendorst (2022) and has incorporated recent advances in natural language processing. Unlike Structural Topic Models, BERTopic is not a probabilistic generative model. Instead, it is an embedding-based approach that combines document representations, dimensionality reduction, and clustering to derive topics.

A central component of BERTopic is the use of word embeddings, which represent the meaning of words or documents as vectors of numbers (Mikolov et al., 2013). Each dimension in the vector can be understood as capturing some aspect of meaning, although the exact interpretation of each dimension is typically not directly interpretable (Figure 2). There are various embedding models that differ in their training data, model architecture, and dimensionality. For example, BERTopic's default embedding model is all-MiniLM-L6-v2 sentence-transformer model (Wang et al., 2020), which has 384 dimensions, while other models, such as OpenAI's text-embedding-3-small (OpenAI, 2024), produce higher-dimensional representations (1,536 dimensions). Higher-dimensional representations can capture more complex patterns in the data and allow for more nuanced representations of word meaning.

**Figure 2**

*Simplified Illustration of Word Embeddings*

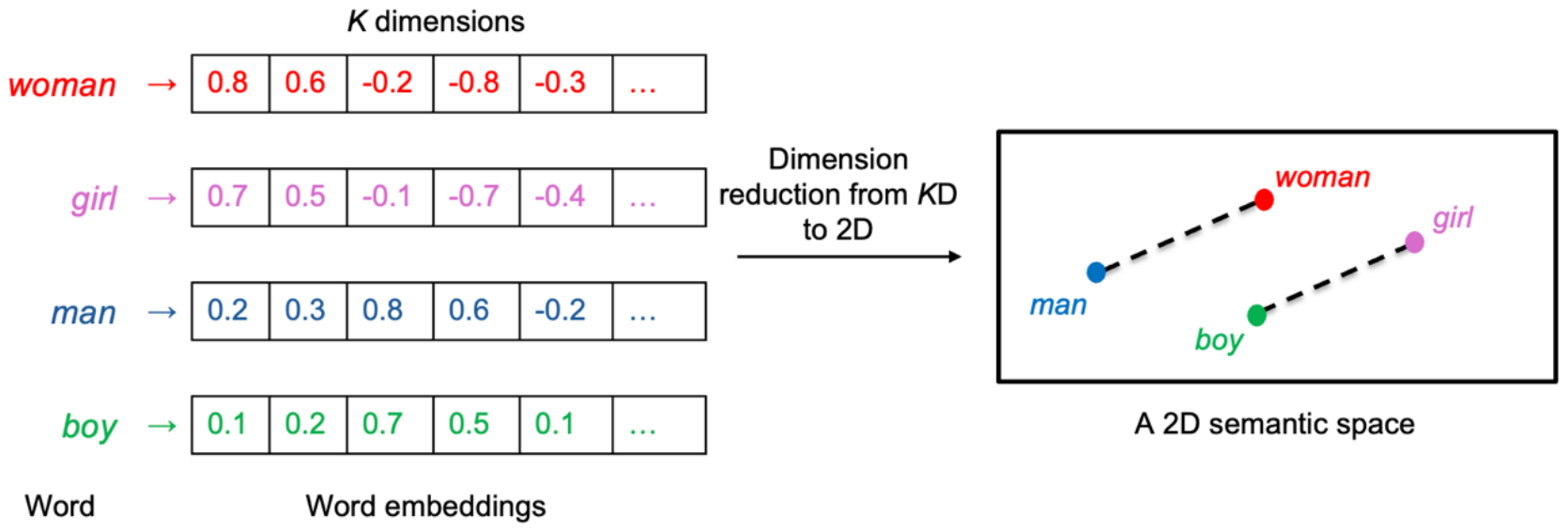

The typical workflow of BERTopic is illustrated in Figure 3. First, each document in the corpus is transformed into a contextualized embedding vector using a pretrained language model. Because distances in very high-dimensional spaces can become less informative, BERTopic next applies dimensionality reduction, most commonly using UMAP (McInnes et al., 2018), to project the embeddings into a lower-dimensional space while preserving their relative structure. After dimensionality reduction, BERTopic applies a clustering algorithm, typically HDBSCAN (McInnes et al., 2017), to group similar documents. HDBSCAN identifies clusters based on regions of high density in the embedding space, without requiring the number of clusters to be specified in advance. Documents that do not clearly belong to any dense region are labeled as outliers. Because clustering assigns each document to a single cluster, BERTopic can be characterized as a single-membership approach, in contrast to mixed-membership models such as LDA or STM.

Once clusters are identified, BERTopic constructs topic representations, defined as sets of topic words generated using class-based TF-IDF (c-TF-IDF; Grootendorst, 2022). In this step, all documents within a cluster are treated as a single combined document, and words that are frequent within the cluster but relatively rare across other clusters are given higher importance (Grootendorst, 2022). This allows the model to extract representative keywords for each topic. Additional optional steps can be used to refine topic representation and improve interpretability.

**Figure 3**

*Typical Workflow of BERTopic*

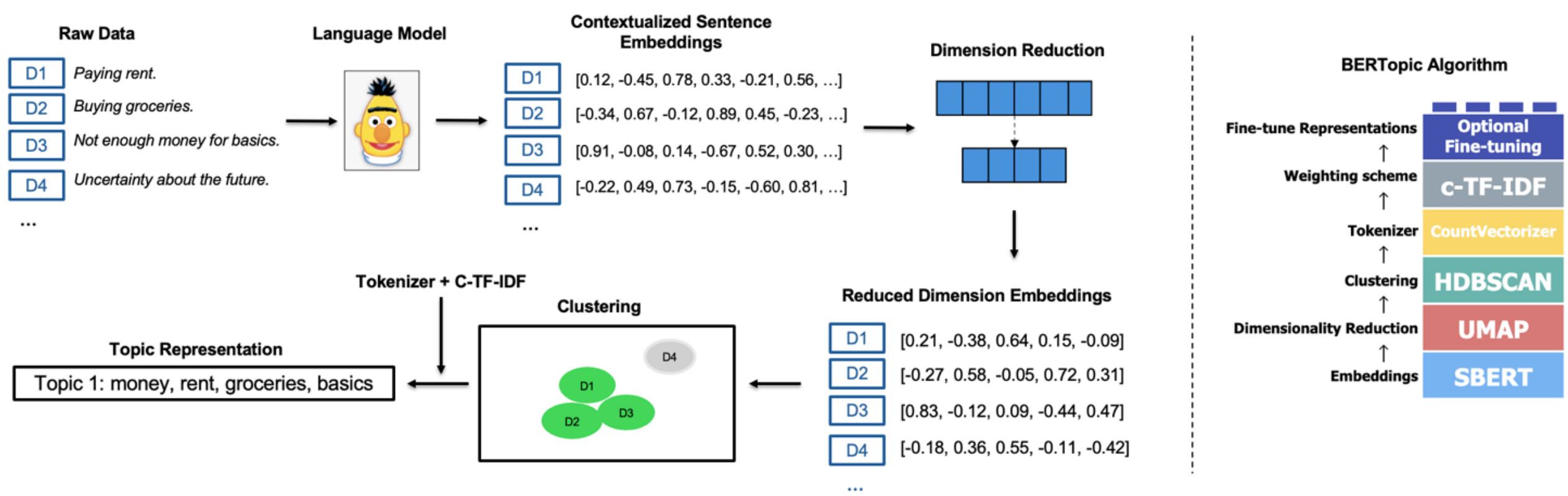

*Notes.* Based on the BERTopic framework described in Grootendorst (2022). Clusters are formed in the embedding space, while topic representations are generated through c-TF-IDF on the original documents using a bag-of-words representation within each cluster.

An important feature of BERTopic is its modular design (Grootendorst, 2022). Each component of the pipeline, including the embedding model, dimensionality reduction method, and clustering algorithm, can be modified independently. For example, researchers may choose different embedding models with varying dimensionality, replace UMAP with alternative reduction techniques such as principal component analysis, or use different clustering algorithms. This flexibility allows BERTopic to adapt to different types of data and to incorporate ongoing developments in natural language processing. In the present study, we focus on the default configuration, which reflects common practice.

Although BERTopic is relatively recent, its use has grown rapidly across both empirical studies and meta-scientific analyses. For example, Cheng et al. (2026) applied BERTopic to analyze open-ended responses in which U.S. participants described metaphors for AI, identifying major themes and examining how these perceptions relate to willingness to adopt AI. Similarly, Newall et al. (2026) applied BERTopic to analyze free-text responses from U.K. gamblers in an online experiment, identifying themes in how participants perceived and reacted to positive emotional gambling harm prevention messages. In addition to empirical applications, BERTopic has been used in bibliometric and review studies (e.g., Fan et al., 2024).

**STM and BERTopic in Open-Ended Survey Responses**

Previous studies have compared LDA-based models with BERTopic, primarily focusing on their ability to identify coherent topics in large text corpora. For example, Grootendorst (2022) compared BERTopic with LDA in the context of news article analysis and reported that BERTopic produced higher-quality topics. However, in that evaluation, documents with fewer than five words were excluded, reflecting a common practice in model development but one that may limit relevance for applied settings where short texts are prevalent. Similarly, Egger and Yu (2022) compared BERTopic with LDA and other topic models in the analysis of Twitter data, highlighting the strengths of BERTopic in capturing themes in short-form social media content. However, their comparison was largely qualitative, and Twitter data differ in important ways from open-ended survey responses, which often include extremely short responses.

Moreover, existing comparisons have focused on topic discovery and have rarely included STM as a central point of comparison. While STM is reduced to an implementation of correlated topic models when covariates are not included, its primary value in social science research lies in its ability to integrate topic estimation with covariates analysis (Roberts et al., 2014). Consequently, there remains a gap in understanding how STM and BERTopic compare in applied settings.

In the present study, we extend prior work by comparing STM and BERTopic using an applied dataset of open-ended survey responses from early care and education providers, while also examining how variations in data preprocessing and model specifications influence model performance. Specifically, we compare model performance before and after text cleaning. We also examine whether stemming improves STM performance, given ongoing debates in the literature regarding its impact (Schofield & Mimno, 2016; Weston et al., 2023). For BERTopic, we evaluate the use of different embedding models. In addition, we introduce contextual augmentation for short responses to examine whether incorporating additional context improves embedding-based model performance.

# Method

## Data

This study draws on data from a national survey of early care and education (ECE) providers in the United States, conducted since March 2021 (approved by the University

Institutional Review Board; Authors, 2024). Participants completed an online baseline survey followed by ongoing follow-up surveys, which were administered biweekly during 2021 and monthly thereafter. The focal question for this analysis was: “What are the biggest challenges and concerns for you and your family right now?” Responses to this question were collected across 127 survey waves. The current analytic corpus includes 14,238 responses from 4,873 providers who reported challenges from March 2021 to October 2025 (see Table 2 for demographics).

**Table 2**

*Sample Demographic Characteristics*

| Characteristics | Category | n | % |
|---|---|---|---|
| Gender | Female | 4,264 | 87.5 |
| | Male | 521 | 10.7 |
| | Non-binary | 17 | 0.3 |
| | Other | 35 | 0.7 |
| | Missing | 36 | 0.7 |
| Provider type | Center-based | 3,011 | 61.8 |
| | Home-based | 1,715 | 35.2 |
| | Missing | 147 | 3.0 |
| Poverty | Below 200% FPL | 1,426 | 29.3 |
| | Between 200–400% FPL | 1,579 | 32.4 |
| | Above 400% FPL | 1,024 | 21.0 |
| | Missing | 844 | 17.3 |
| Race/ethnicity | White | 3,155 | 64.7 |
| | Black | 683 | 14.0 |
| | Latinx | 680 | 14.0 |
| | Other racial and ethnic groups | 314 | 6.4 |
| | Missing | 41 | 0.8 |
| Education | High school or equivalent or less | 390 | 8.0 |
| | Some college or Associate’s | 1,737 | 35.6 |
| | Bachelor’s degree or higher | 2,417 | 49.6 |
| | Missing | 329 | 6.8 |

Responses were highly variable in length, averaging 15.6 words (range = 1-299). Short responses were common: 7.4% consisted of a single word, and 34.6% contained between one and five words. The raw text also contained various forms of noise, including encoding-related artifacts (e.g., “,Äî” instead of dashes) and typographical errors. We identified 1,163 typographical errors that were likely to affect topic modeling results (e.g., “Uncertanty” versus

“Uncertainty”), particularly in very short responses where each word carries substantial weight. To examine the impact of such data conditions, we constructed both raw and cleaned versions of the corpus, with cleaning primarily addressing typographical errors.

## Data Analysis

### *Model Setup*

We compared the performance of STM and BERTopic under a set of conditions. These conditions were designed to examine how preprocessing decisions and modeling strategies influence topic modeling performance for short, open-ended survey responses (Table 3).

**STM Conditions.** We estimated STM under three preprocessing conditions. First, we fit STM on the original corpus without correcting typographical errors (“raw”). Second, we applied typographical correction to the text prior to modeling (“clean”), addressing spelling errors that could affect word frequency and topic estimation. Third, we applied stemming to the cleaned text (“clean+stem”), reducing words to their root forms (e.g., “working” to “work”). The use of stemming in topic modeling remains debated (Schofield & Mimno, 2016; Weston et al., 2023), as it may change the word frequency structure in bag-of-words representations while reducing interpretability; we therefore include it as a separate condition.

**BERTopic Conditions.** We evaluated BERTopic under five conditions. First, we applied BERTopic to the original corpus without correcting typographical errors (“raw”). Second, we applied BERTopic to text in which typographical errors were corrected prior to modeling (“clean”). Third, we examined the effect of embedding choice by replacing the default embedding model (*all-MiniLM-L6-v2*, 384 dimensions) with a higher-dimensional embedding model (*text-embedding-3-small*, 1,536 dimensions; “clean+emb”). We include this comparison because higher-dimensional embedding models are often trained on larger and more diverse datasets and may capture more nuanced semantic relationships in the data, which could be beneficial for analyzing short texts. Fourth, we introduced contextual augmentation by appending the survey question to each response (see an example in Table 3), with the goal of providing additional semantic context for very short responses ("clean+ctx"). Because BERTopic relies on sentence embeddings rather than a bag-of-words representation for document clustering, adding this context may help the model better cluster documents without directly altering the word frequency distributions used for topic representation. Finally, we combined contextual augmentation with the higher-dimensional embedding model to examine their joint effects ("clean+ctx+emb").

**Table 3**

*Summary of Model Conditions*

| Condition | Description |
|---|---|
| STM: raw | Original text without correcting typographical errors |
| STM: clean | Text with typographical errors corrected |
| STM: clean+stem | Cleaned text with stemming applied to reduce words to root forms |
| BER: raw | Original text without typographical correction |
| BER: clean | Text with typographical errors corrected |
| BER: clean+emb | Cleaned text with higher-dimensional embedding model (*text-embedding-3-small*, 1,536 dimensions) |
| BER: clean+ctx | Survey question appended to each cleaned response to provide additional semantic context, e.g.:<br>● Raw response: "Income."<br>● Added context: "Question: What are the biggest challenges and concerns for you and your family right now? Answer: Income." |
| BER: clean+ctx+emb | Context-augmented text with higher-dimensional embedding model |

**Implementation.** To reduce sensitivity to a single choice of topic number, all model conditions were estimated across a range of topic numbers ($K$ = 10, 20, 30, 40, 50, 60) and with three random initializations (seeds = 42, 66, 88) (Angelov, 2020; Grootendorst, 2022). This resulted in 18 model runs per condition and a total of 144 model runs across all conditions. BERTopic models were implemented in Python using Google Colab (Grootendorst, 2022), whereas STM models were estimated using the stm package in R (Roberts et al., 2019). For STM, preprocessing was conducted using its built-in textProcessor function, which applies procedures including lowercasing, removal of punctuation and stop words, and filtering of extremely rare and frequent terms. For BERTopic, which does not require preprocessing for document representation by design, we applied similar preprocessing steps, including lowercasing and removal of punctuation and stop words, to facilitate a fair comparison across models.

For BERTopic conditions involving contextual augmentation, the augmented text was used directly for embedding extraction without preprocessing. This allows the embedding model to utilize the full context of both the appended question and the response when generating document embeddings. The preprocessed text was used for topic representation. This approach reflects the two-stage nature of BERTopic, in which document clustering based on embeddings and topic representation based on c-TF-IDF are separate steps, allowing flexibility in how text is processed at each stage.

### *Covariates Analysis*

We conducted covariate analyses for both STM and BERTopic. Two covariates were examined, including time (continuous variable) and provider type (categorical variable). Time was recorded at the year–month level based on when each survey response was collected. Provider type was coded as center-based and home-based. In STM, covariates are incorporated during model estimation, whereas in BERTopic, covariate differences are examined after topics are generated, providing descriptive rather than inferential analyses.

### *Evaluation Methods*

Both quantitative and qualitative methods were used to evaluate model performance across models and conditions.

We employed two commonly used quantitative metrics to assess topic quality: topic coherence and topic diversity. Specifically, we used normalized pointwise mutual information (NPMI; Bouma, 2009) to evaluate topic coherence, which has been shown to correlate with human judgements (Lau et al., 2014). NPMI measures the extent to which pairs of words within a topic co-occur in the corpus relative to chance, based on the assumption that coherent topics consist of words that frequently appear together. NPMI ranges from −1 to 1, with higher values indicating greater topic coherence.

To assess topic diversity, we used Inverted Rank-Biased Overlap (Inverted RBO; Webber et al., 2010). Inverted RBO evaluates the distinctiveness of topics by comparing the ranked lists of top words across topics, placing greater weight on overlap among higher-ranked words. The measure ranges from 0 to 1, with higher values indicating greater topic diversity. Both metrics were computed using the OCTIS Python package (Terragni et al., 2021).

In addition to quantitative metrics, we conducted qualitative evaluations of topic interpretability across models and conditions. We also examined model stability and alignment by assessing overlap at both the word and topic levels. Specifically, we assessed word-level overlap using the proportion of shared words and topic-level overlap using the number of shared topic words between topic pairs. In addition, because both models exclude portions of the data during analysis, we examined data retention rates across conditions.

# Results

## Quantitative Results

Across all conditions, BERTopic consistently performed better than STM in topic coherence (Figure 4). On average, STM produced negative NPMI values across conditions (raw: −0.08, clean: −0.08, clean+stem: −0.07), indicating modest co-occurrence among topic words, whereas BERTopic achieved substantially higher coherence (raw: 0.09, clean: 0.09). Among BERTopic conditions, contextual augmentation led to the largest improvements, with clean+ctx (0.12) and clean+ctx+emb (0.11) showing the highest coherence scores. In contrast, using higher-dimensional embeddings alone (clean+emb: 0.07) did not improve performance and in fact reduced coherence relative to the default embedding.

**Figure 4**

*Topic Coherence across Conditions and Models*

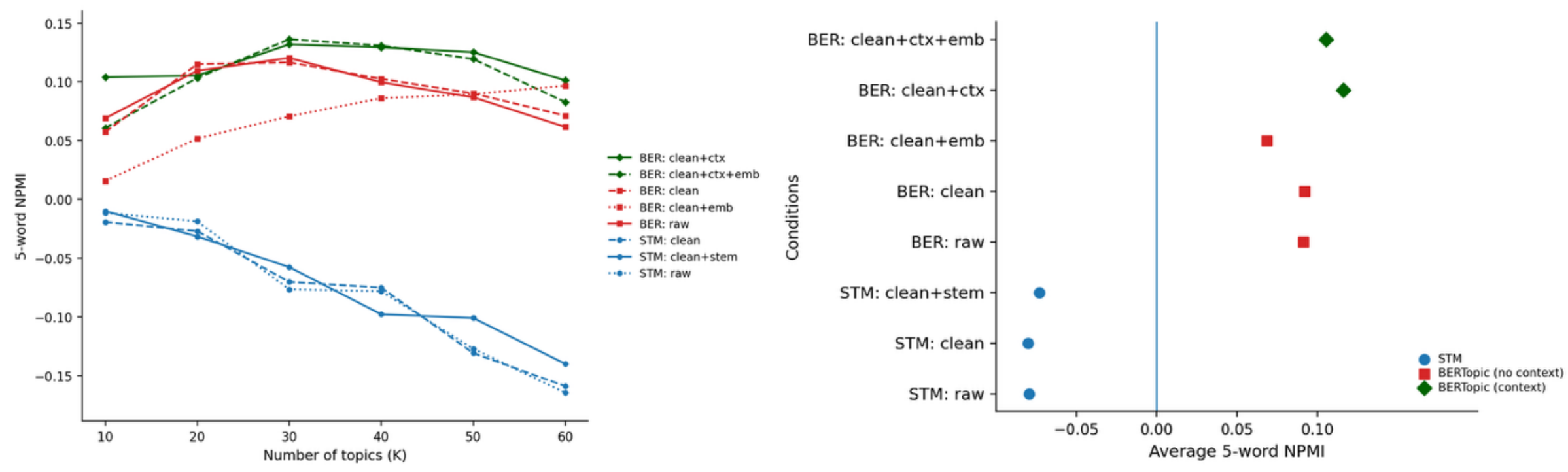

In terms of topic diversity, STM exhibited near-perfect scores across all conditions (Figure 5), while BERTopic showed slightly lower but still high diversity (0.98-0.99).

**Figure 5**

*Topic Diversity across Conditions and Models*

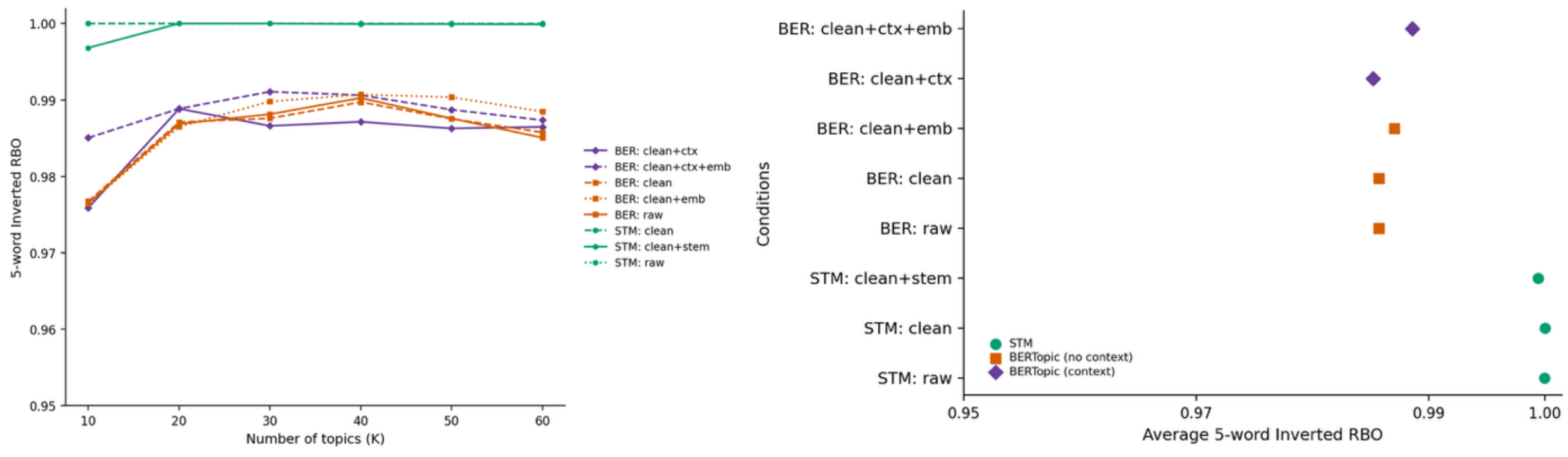

In terms of data retention, STM removes low- and high-frequency words during preprocessing, whereas BERTopic assigns some documents to an outlier category and excludes them from topic formation. In STM, 51.9%–57.9% of words were retained, whereas in BERTopic, 59.4%–72.7% of documents were retained. Overall, cleaner data resulted in higher retention rates across conditions (Figure 6). However, using higher-dimensional embeddings with limited contextual information was associated with greater data loss than using the default embedding model in our analysis.

**Figure 6**

*Data Inclusion Rate by Condition*

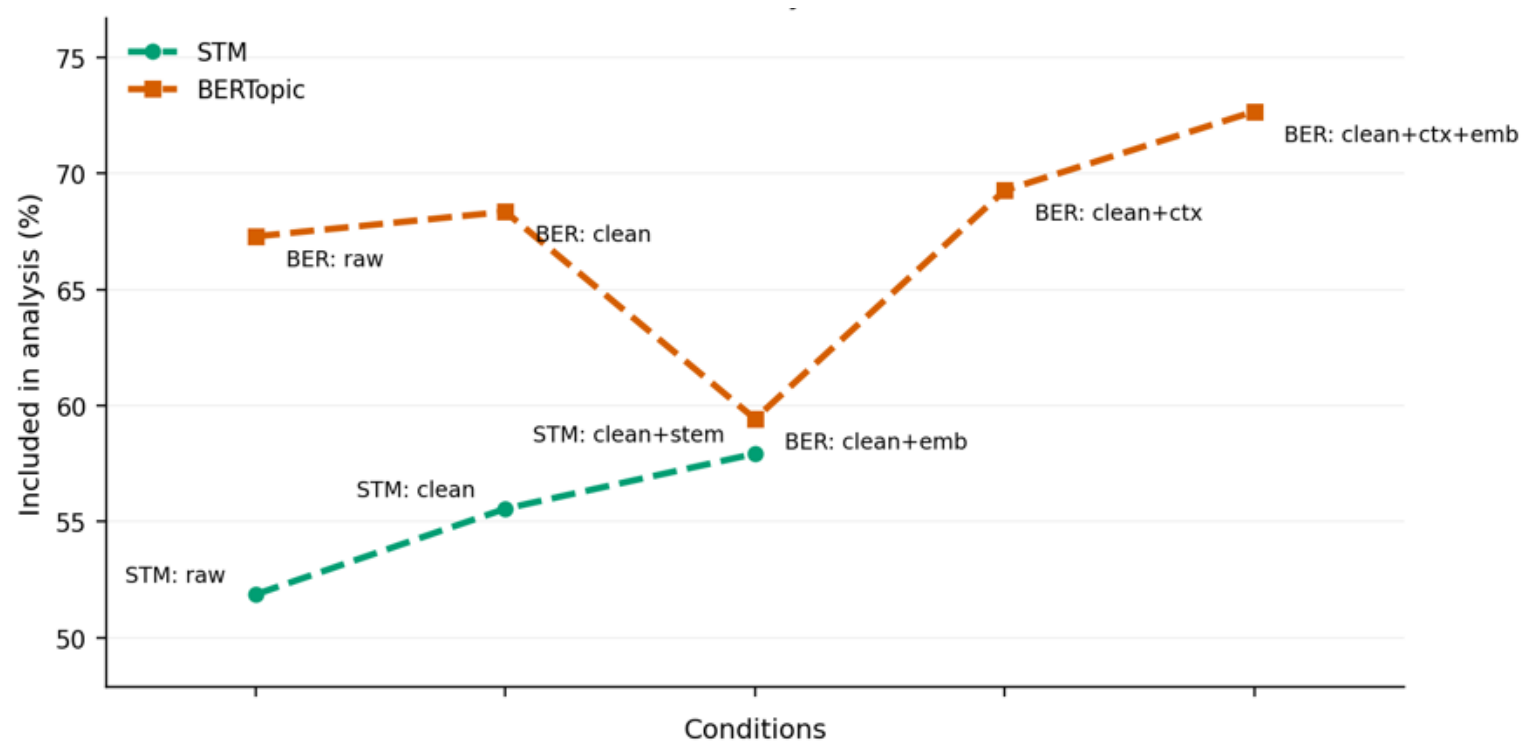

*Note.* Inclusion rate is defined differently across models: for STM, it reflects the proportion of words retained after preprocessing, whereas for BERTopic, it reflects the proportion of documents assigned to topics.

## Qualitative Results

Given the large number of models estimated across seeds and topic numbers, qualitative results reported below are based on conditions and models with a fixed seed (42) and number of topics ($K$ = 30). Table 4 presents the top five words for the most prevalent topics across eight model conditions. In terms of topic interpretability. Across all STM conditions, topics tended to be less distinct and often reflected mixtures of multiple themes. For example, a topic in the raw condition with words such as “financial,” “covid,” “dont,” “finding,” and “government” appears to combine themes related to COVID, financial hardship, and government concerns. In addition,

STM topics sometimes included general terms (e.g., "cant," "still," "people," "just"), which made interpretation less straightforward. In contrast, BERTopic produced more clearly defined and interpretable topics across all conditions, such as basic needs (e.g., "food," "gas," "groceries"), financial stress (e.g., "bills," "debt," "credit"), housing (e.g., "rent," "housing"), and health (e.g., "insurance," "medical"). Compared with STM, BERTopic topics exhibited sharper boundaries between themes, with less mixing of concepts.

**Table 4**

*Top Five Words of the Most Prevalent Topics across Models and Conditions*

| **STM: raw** | | | | | | | |
|---|---|---|---|---|---|---|---|
| healthy | money | cost | financial | care | children | pay | health |
| costs | enough | living | covid | biggest | future | insurance | work |
| making | just | mental | dont | challenge | still | parents | quality |
| rising | trying | high | finding | like | lack | medical | two |
| cant | make | paid | government | even | people | new | days |
| **STM: clean** | | | | | | | |
| cost | health | family | care | time | healthy | able | work |
| food | pay | children | bills | take | staying | everything | school |
| financial | enough | keeping | will | sure | new | kids | husband |
| living | don't | biggest | parents | worried | people | families | support |
| just | insurance | live | way | state | day | rent | quality |
| **STM: clean+stem** | | | | | | | |
| cost | pay | food | money | stress | need | famili | get |
| time | husband | stay | keep | worri | afford | job | school |
| care | debt | healthi | year | dont | hous | provid | month |
| enough | cant | futur | medic | just | take | member | want |
| balanc | loan | back | move | rise | basic | social | good |
| **BER: raw** | | | | | | | |
| childcare | financial | food | bills | housing | insurance | covid | healthy |
| children | finance | gas | debt | rent | medical | pandemic | staying |
| child | stability | groceries | paying | house | health | getting | health |
| care | expenses | utilities | credit | home | healthcare | virus | safe |
| daycare | family | cost | card | repairs | bills | exposure | everyone |
| **BERT: clean** | | | | | | | |
| childcare | financial | food | insurance | housing | bills | covid | healthy |
| child | finances | gas | medical | rent | debt | pandemic | staying |
| care | stability | groceries | health | house | paying | getting | health |
| children | expenses | utilities | healthcare | repairs | credit | free | everyone |
| daycare | family | cost | bills | home | card | exposure | healthcare |
| **BERT: clean+emb** | | | | | | | |
| childcare | cost | covid | financial | health | husband | paying | healthy |
| child | rising | vaccinated | finances | medical | retirement | bills | staying |
| care | prices | vaccine | stability | insurance | spouse | rent | safe |
| children | costs | getting | stress | healthcare | retire | pay | keeping |
| daycare | food | virus | challenges | care | job | bill | everyone |
| **BERT: clean+ctx** | | | | | | | |
| financial | food | time | health | childcare | covid | bills | work |
| finances | cost | children | insurance | daycare | getting | debt | job |
| money | living | together | medical | child | free | paying | balance |
| income | prices | family | healthcare | care | get | credit | hours |
| ends | rising | stress | issues | center | exposure | card | life |
| **BERT: clean+ctx+emb** | | | | | | | |
| bills | childcare | cost | health | covid | job | healthy | mental |
| debt | care | inflation | insurance | vaccinated | economy | staying | stress |
| money | child | rising | medical | vaccine | employment | safe | health |
| financial | daycare | living | healthcare | virus | economic | keeping | depression |
| paying | children | prices | care | getting | husband | everyone | anxiety |

To assess topic stability and overlap within and across models, we compared results from the raw and clean conditions, which use relatively matched preprocessing settings. Within-model comparisons indicate that BERTopic produces more stable topics across preprocessing conditions than STM (Table 5). Specifically, the overlap between BERTopic raw and clean conditions was high (86.7%), with many topics sharing four or five words, indicating high consistency in topic structure. In contrast, STM raw and clean conditions showed lower overlap (78.7%), with most topics sharing only one or two words.

Across models, topic overlap between STM and BERTopic was modest. Word overlap was slightly above 40% under both raw and clean conditions, and no topics shared more than two words. This relatively low overlap can be partly explained by differences in how the two models represent topics. STM topics often include more general terms (e.g., “cant,” “just”), whereas BERTopic tends to produce more specific and semantically consistent terms. In addition, BERTopic captures variations of related words (e.g., “vaccine,” “vaccination,” “vaccinated”), while STM is less likely to group these variants. STM topics also more frequently reflect mixtures of multiple themes, which further reduces direct topic-level alignment with BERTopic topics. Despite these differences, both models capture similar high-level themes, including financial strain, basic needs, housing, debt, COVID-related concerns, physical and mental health, and family responsibilities.

**Table 5**

*Within- and Cross-Model Word- and Topic-Level Overlap*

| Condition | Word overlap (%) | Topic overlap (count) | | | | | |
|---|---|---|---|---|---|---|---|
| | | 5 words | 4 words | 3 words | 2 words | 1 word | 0 word |
| STM: raw vs clean | 78.7 | 0 | 0 | 0 | 8 | 18 | 4 |
| BERT: raw vs clean | 86.7 | 18 | 9 | 0 | 1 | 0 | 2 |
| Raw: STM vs BERT | 40.7 | 0 | 0 | 0 | 5 | 16 | 9 |
| Clean: STM vs BERT | 42.7 | 0 | 0 | 0 | 6 | 15 | 9 |

## Covariates Analysis Results

We conducted covariate analyses for both models, focusing on time and provider type. These analyses are illustrative and were based on STM (clean+stem) and BERTopic (clean+ctx+emb) conditions, as these retained the most data. We examined two topics identified in both models: high cost of living and health insurance. Again, STM incorporates covariates during the topic estimation process, enabling inferential analysis that accounts for uncertainty and provides estimates with associated confidence (Roberts et al., 2014). In contrast, covariate analysis in BERTopic is conducted post hoc and remains descriptive, as illustrated in Figure 7.

**Figure 7**

*Topic Prevalence by Time and Provider Type*

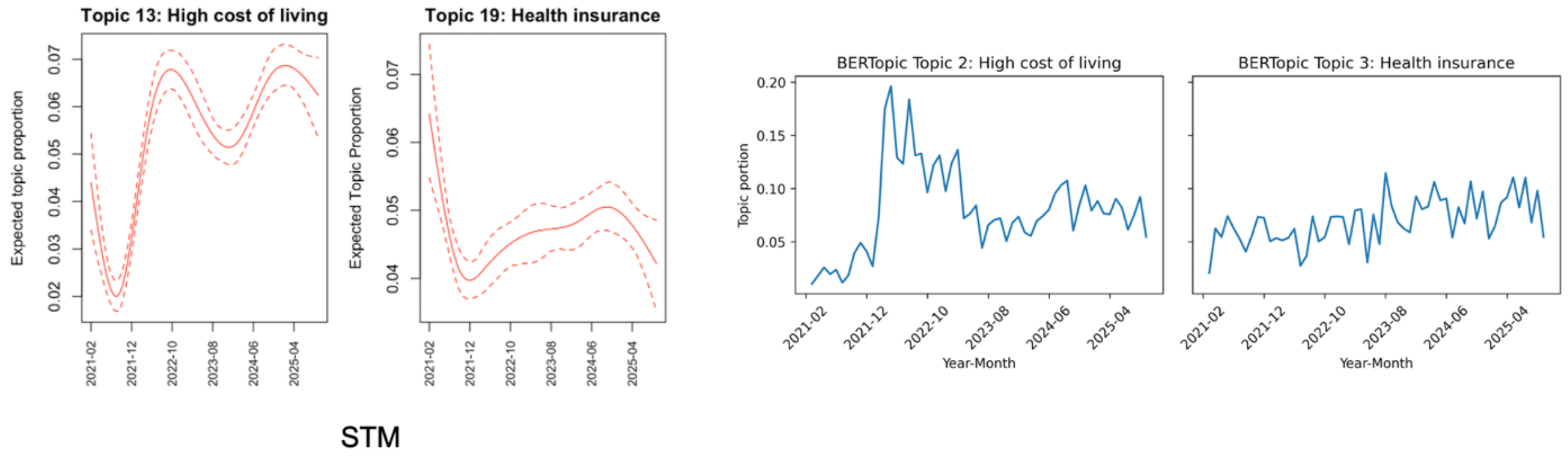

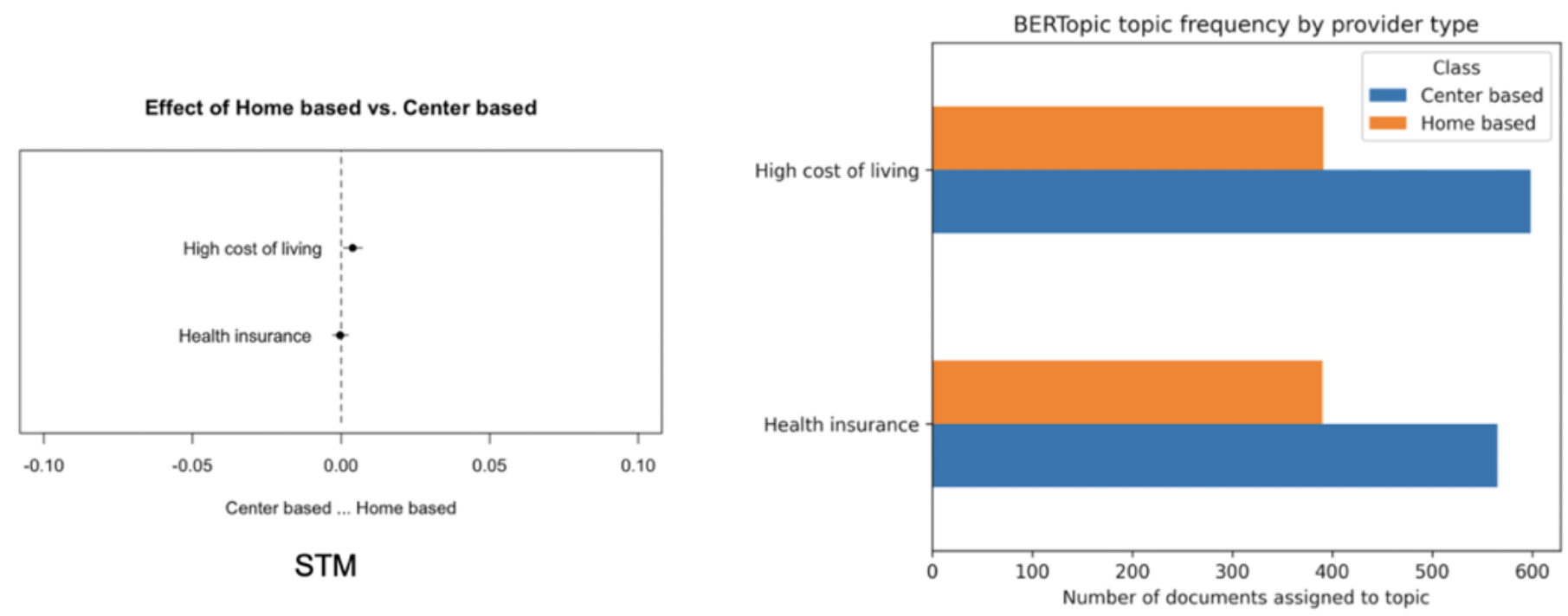

*Note*. Topic prevalence is measured as proportions rather than raw frequencies; therefore, temporal patterns should be interpreted within each model's own set of identified topics.

## Discussion

This study highlights that model selection and specification play important roles in topic modeling for short, open-ended survey responses, a common but underexamined data setting in social science research. Our findings show that model performance in such contexts cannot be assumed to mirror results from benchmark corpora (Laureate et al., 2023). Instead, researchers need to carefully consider both corpus characteristics and model behavior. Document characteristics, preprocessing decisions, and model setups can all influence results. These findings underscore the importance of researchers' familiarity with the data (Grimmer et al., 2013). Even when computational methods are used to scale text analysis, researchers should actively engage with their data, including reading responses and making informed preprocessing decisions. Rather than relying on standardized pipelines, data preparation in applied settings requires context-specific judgment and problem solving.

Across models, BERTopic generally produced more coherent and interpretable topics and exhibited greater stability across preprocessing conditions, consistent with previous studies (Egger & Yu, 2022), making it particularly useful for topic discovery and exploratory analysis. However, its outputs are primarily descriptive. In contrast, STM provides a framework for incorporating covariates during model estimation, supporting inferential analysis of how topics vary across time and groups. At the same time, our results suggest that the quality of identified topics should be carefully evaluated before conducting downstream analyses with STM, as lower coherence and mixed-topic representations may affect interpretability. These findings point to a complementary relationship between the two approaches: BERTopic may be better suited for identifying meaningful topic structures, while STM offers advantages for hypothesis testing and inference once topic quality is established.

Our results also provide insight into how modeling choices affect performance in short-text settings. Increasing embedding dimensionality alone did not improve results and, in some cases, was associated with reduced performance, suggesting that more complex representations are not necessarily beneficial for sparse data. In contrast, adding contextual information substantially improved the performance of embedding-based models, highlighting the importance of input representation when responses are short and lack sufficient standalone context. The modular structure of BERTopic allows flexibility at different stages of the modeling pipeline, enabling researchers to tailor representation, clustering, and topic construction to the needs of the data (Grootendorst, 2022).

Finally, our findings underscore the limitations of relying solely on quantitative evaluation metrics (Dieng et al., 2020). Measures such as topic coherence and topic diversity capture important aspects of model performance but might not fully reflect topic quality (Grootendorst, 2022). For example, topics containing frequent but generic words may achieve high coherence scores while remaining difficult to interpret (Roberts et al., 2014), and topics with distinct but uninformative words may appear diverse. As a result, combining quantitative metrics with qualitative assessment is essential for meaningful evaluation. More broadly, topic modeling should be understood as a semi-automated process that requires human judgment (Grimmer et al., 2013; Roberts et al., 2014). Model outputs might not represent a single, objective structure inherent in the data; rather, they reflect analytic choices made by the researcher (Roberts et al., 2014). As such, meaningful interpretation requires domain knowledge, and results should be evaluated in light of both research goals and qualitative assessment alongside quantitative metrics.

Based on insights from our analyses, we offer practical recommendations for researchers using these two models. Although many topic modeling approaches exist, we focus on STM and BERTopic as representative examples, and similar considerations may apply to other models. Figure 8 presents a framework to guide the selection and application of topic modeling approaches based on primary research goals. The figure emphasizes that model choice should be driven by analytic purpose rather than default preference. For topic discovery, BERTopic could be advantageous due to its ability to generate more coherent and interpretable topics, especially for short and noisy text; however, STM remains a viable alternative depending on data characteristics. For inference with covariates, STM is recommended because it incorporates covariates directly into the modeling process, enabling formal statistical analysis. For mixed goals that involve both discovery and inference, we suggest an iterative workflow: researchers may first use BERTopic to discover topics and use the discovered topic structure to inform the following STM inferential analysis. Across all scenarios, the framework highlights the importance of evaluating topic quality, iterating on preprocessing and modeling decisions, and aligning analytic choices with research goals.

**Figure 8**

*Guiding Framework for Using STM and BERTopic in Survey Data Analysis*

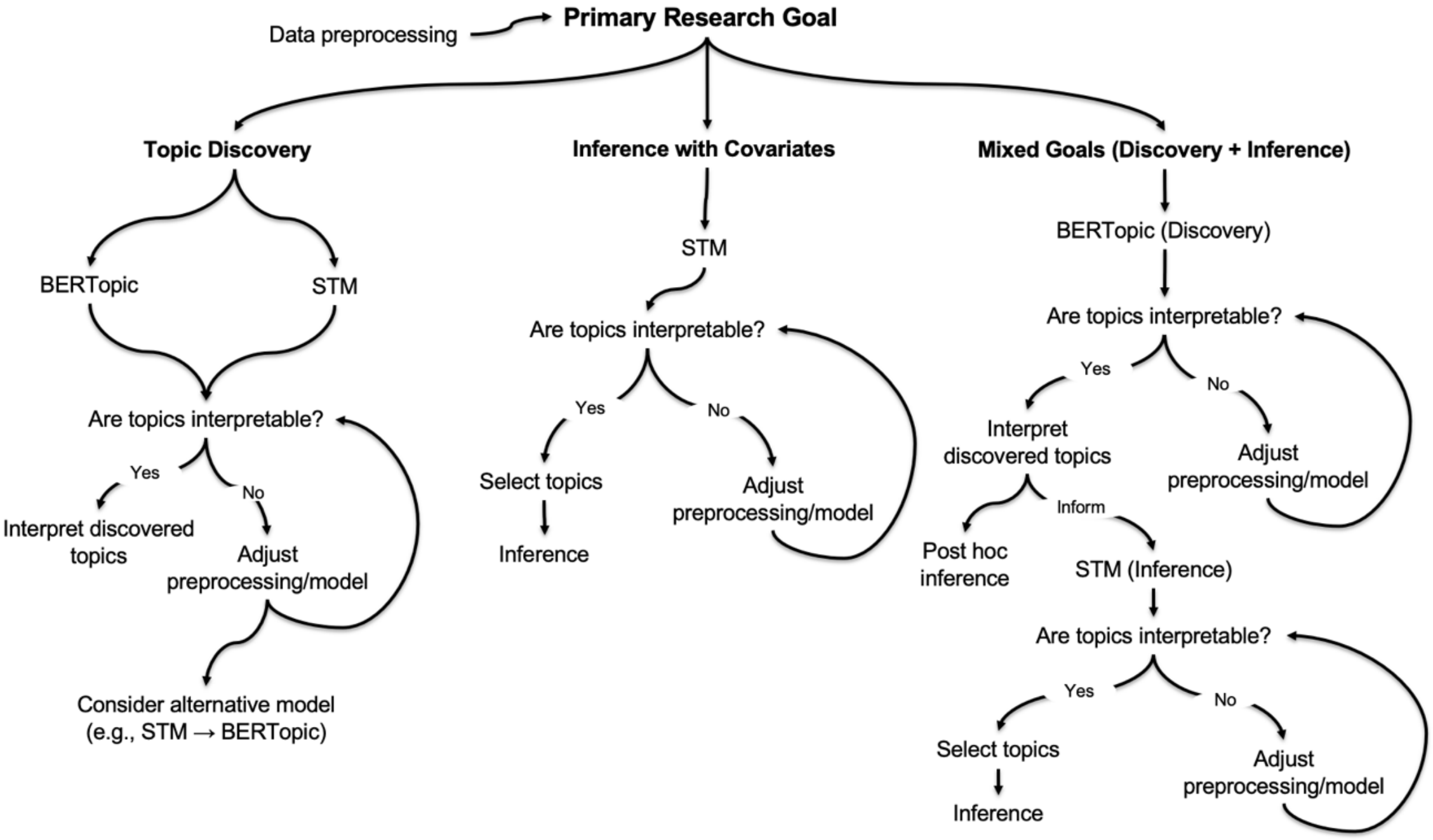

Several limitations should be kept in mind when interpreting the results. First, our analyses are based on short, open-ended survey responses, and the findings may not generalize directly to other types of text data. Although our results could provide some useful insights, separate evaluations are needed for these other data contexts. Second, we focused on relatively basic implementations of both models. For STM, more advanced specifications (e.g., "Spectral" initiation) were not explored, and for BERTopic, we relied primarily on its default pipeline despite its flexibility and many possible configurations. In addition, we did not attempt to identify an optimal number of topics, and therefore the reported results should not be interpreted as definitive representations of the corpus. Third, both models include additional functionalities that may be valuable for researchers but were beyond the scope of this study. For example, BERTopic supports hierarchical topic structures and multilingual analysis, and its modular architecture allows integration of newly developed techniques. Future work could explore these features more systematically to better understand their implications for applied research.

## Conclusion

In this study, we compared STM and BERTopic in the context of short, open-ended survey responses, a common but underexamined data setting in social science research. Our findings highlight that model performance depends not only on the method itself but also on data characteristics, preprocessing decisions, and analytic goals. Rather than viewing these approaches as competing alternatives, we suggest that STM and BERTopic offer complementary strengths for different stages of analysis. We hope this work provides practical guidance for researchers and encourages more thoughtful, context-sensitive applications of topic modeling in applied research.